# FAKER: Full-body Anonymization with Human Keypoint Extraction for Real-time Video Deidentification

Byunghyun Ban[*], Hyoseok Lee
Nana Co., LTD.
https://apicloud.ai
{bhban, hslee}@apicloud.ai

**Abstract**

*In the contemporary digital era, protection of personal information has become a paramount issue. The exponential growth of the media industry has heightened concerns regarding the anonymization of individuals captured in video footage. Traditional methods, such as blurring or pixelation, are commonly employed, while recent advancements have introduced generative adversarial networks (GANs) to redraw faces in videos. In this study, we propose a novel approach that employs a significantly smaller model to achieve real-time full-body anonymization of individuals in videos. Unlike conventional techniques that often fail to effectively remove personal identification information such as skin color, clothing, accessories, and body shape while our method successfully eradicates all such details. Furthermore, by leveraging pose estimation algorithms, our approach accurately represents information regarding individuals' positions, movements, and postures. This algorithm can be seamlessly integrated into CCTV or IP camera systems installed in various industrial settings, functioning in real-time and thus facilitating the widespread adoption of full-body anonymization technology.*

▪*Keyword : video anonymization, privacy protection, full-body de-identification, video processing, computer vision*

## 1. Introduction

### 1.1. Background

In modern society, the protection of personal information is becoming an increasingly critical issue [1]. With the explosive growth of video content, the need for technologies that safeguard personal identities has been emphasized. Videos captured in public places, schools, workplaces, and other settings often expose individuals' faces or bodies, which can lead to privacy violations. To address this issue, there is a need for anonymization technologies that can remove individuals from videos.

### 1.2. Problem Statement

Recent approaches tire to recognize and regenerate new face with generative AI models for anonymization [2-6] and traditional technologies primarily rely on blurring or pixelating faces [8]. However, these methods can degrade the overall quality of the video and may not effectively protect personal information in certain situations. For example, elements such as clothing, skin color, body shape, gender, and hairstyle can still provide clues to identify individuals. Therefore, a more sophisticated and effective anonymization method is necessary.

### 1.3. Motivation and Contributions

In this study, we propose a method that removes individuals from video frames and then superimposes the estimated skeletal structure using a pose estimation algorithm, specifically BlazePose. This method offers the following advantages:

#### 1.3.1. Removal of Personal Identification

By removing not only faces but entire bodies, the possibility of personal identification is minimized. This approach effectively eliminates elements such as clothing, skin color, body shape, gender, and hairstyle to enhance anonymity.

#### 1.3.2. Minimization of Video Distortion

Unlike methods that add blurring or mosaic effects, our approach avoids adding any such artifacts. Additionally, because we do not use an end-to-end autoencoder-like model that processes entire video frames, the distortion of the original video is minimized.

#### 1.3.3. Provision of Anonymized Videos

Our method retains important information about the presence of individuals, their locations, sizes, poses, and movements within the video. The pose estimation algorithm provides key point locations of the human body, allowing viewers to understand what actions the individuals are performing. This capability is useful not only for monitoring user behavior but also for detecting emergency situations, such as falls or other incidents where individuals are unable to get up. In such emergencies, the system can be configured to send the original video, with user consent, to provide appropriate assistance.

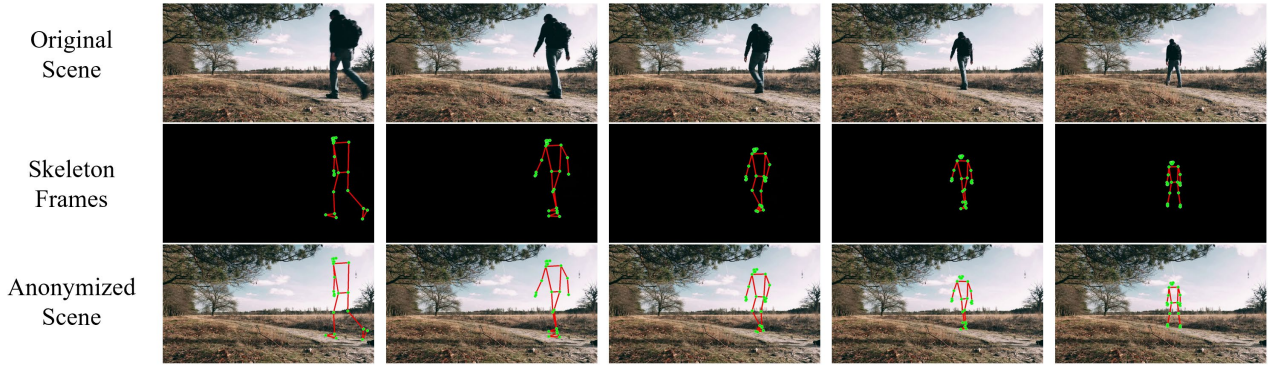

Figure 1. Full Body Anonymization Strategy

### 1.3.4. Smaller Computational Complexity

Previous methods often require extracting human masks using semantic segmentation, removing these regions, and then filling the extracted areas using generative AI, which demands substantial computational resources. Real-time operation would necessitate supercomputers, and using APIs for generative models like DALL-E would introduce response latency and additional API call costs. Instead of using generative AI, we introduce a minimized version of human removal algorithms that rely on non-machine learning-based video processing techniques.

This allows real-time video processing without the need for supercomputers, achieving low-cost real-time anonymization. The reduced computational requirements also enable the method to be deployed on embedded IoT systems, expanding its applicability to various real-world scenarios.

This study introduces a novel methodology that combines Mask R-CNN and BlazePose. Through various experiments, we validate the efficiency and effectiveness of the proposed method, demonstrating its potential for real-time anonymization and its applicability in embedded systems. These contributions aim to advance the field of video anonymization technologies significantly.

## 2. Related Works

### 2.1. Human Anonymization

Human anonymization refers to the process of protecting individual identities from data. Particularly in the field of computer vision processing, algorithms are researched to remove personal information of individuals captured in videos. Common approaches include editing facial images or synthesizing new faces over the original ones. The primary techniques involved in human anonymization include face detection and regeneration.

#### 2.1.1. Face Detection

Face detection is a technology that identifies human faces in videos or images. Various object detection algorithms can be applied, and recently, techniques utilizing YOLO [9] and its variant models have been widely published [10-19]. This process uses machine learning models or deep learning algorithms to locate and determine the size of faces.

Many state-of-the-art human anonymization techniques use object detection algorithms to detect the location and size of human faces in a scene as the first step [2, 3, 5, 7].

#### 2.1.2. Face Regeneration

In the past, anonymization was performed by cropping the detected face regions from the scene and then applying blurring or mosaic effects [2].

However, after the publication of Generative Adversarial Networks (GANs) in 2014 [20], the trend over the past decade has shifted towards using generative AI to create new face images that obscure the original faces beyond recognition. Most research utilizes GANs [2-7], while some researchers have introduced techniques applying Diffusion Models [21]. As it has been shown that adversarial training can contribute not only to the generation of fake images but also to enhancing the quality of general image processing [22-23], variants of GANs are preferred by many researchers because they can generate high-quality fake face images while preserving additional information such as facial angles and expressions [3].

### 2.1.3. Full body anonymization

In 2023, Hukkelås, H., et al. has introduced DeepPrivacy V2 [24] for full body anonymization. Their model first takes an image as input and segments the full bodies of the main individuals to be anonymized. It then attempts full body generation using a continuous surface embedding (CSE) model-guided generator [25], and if the CSE fails in body detection, an unconditional full body generator is employed for body generation. Finally, a face generator is used to alter the faces of the remaining individuals. While this technique effectively regenerates face shapes and clothing, it is incapable of modifying poses or body shapes.

## 2.2. Removal of Human Body form Scene

Our method operates by erasing the entire human body image from the scene and then generating a skeleton image. The task of removing human images in a video can be implemented using various object removal techniques. While removing static objects such as trees or buildings in videos can be achieved through local similarity analysis [26], many researchers use generative AI to redraw the background in areas where humans have been removed.

### 2.2.1. Recognition of Human Body

To remove human bodies from a video and seamlessly reconstruct the background in the erased areas, the first step is to identify the pixels occupied by the human bodies in the original scene. This can be easily achieved using semantic segmentation or object detection algorithms. Semantic segmentation typically involves an autoencoder-like model that compresses and then decompresses the image, performing pixelwise classification during this process [27-30]. On the other hand, object detection generally uses classification on larger-scale patches to achieve this goal [31-33].

### 2.2.2. Background Regeneration

By identifying human regions in videos and using a generative model to seamlessly erase these parts and generate new images, it is possible to remove human bodies. M. Granados et al. introduced an algorithm that defines an energy function for moving objects and processes pixels to remove individuals [34]. This algorithm generates new images by taking the original scene and a segmentation mask as inputs. Similarly, the algorithm by Qiwen Fu et al. [35] operates by predicting the bounding box of the individual to be removed using object detection, and then inputting the masked area within this box into a GAN to generate a new image.

Unlike the other techniques introduced in this section, our algorithm leverages semantic segmentation but does not employ generative AI models. The limitations of existing generative models in terms of performance are evident, and their high computational requirements make them unsuitable for real-time video processing.

## 2.3. Human Pose Estimation

Human pose estimation is a technique used to predict the positions and states of various body parts of humans captured in videos. It typically involves recognizing keypoints of different body parts and analyzing the positions and distances of these points. Commonly used keypoints include major joints such as knees, elbows, and shoulders. When facial expression recognition is also incorporated, keypoints can include areas around the eyes and the edges of the lips.

### 2.3.1. Joint Location Prediction

In the past, techniques generally involved regressing feature information extracted through video analysis to infer the locations of keypoints [36, 37]. However, since 2014, most published papers have utilized CNNs, and it has become common to infer the positions of all keypoints simultaneously in an end-to-end manner [38-41]. Initially, achieving pose estimation using CNNs was recognized as a significant contribution in itself. However, since 2020, the main research trend has shifted towards model lightweighting. The BlazePose algorithm [42], for instance, uses a lightweight CNN model that allows real-time estimation at 30fps even on old Pixel 2 smartphones.

### 2.3.2. Pose Estimation Based Hazard Detection

Since the result of human pose estimation represents the position information of each major joint of the body, it is possible to perform hazard detection by analyzing the positions and movements of these joints [42-45]. For example, a rapid change in the recognized joint positions in a video indicates body movement. If most of the upper body joints detected in the video rapidly descend to the floor, it can be inferred that a fall has occurred. If the heights of most joints are similar and the difference in the y-coordinates between the lowest and highest joints is small, it can be deduced that the person is lying down. Many recent studies have utilized this type of information to implement hazard detection.

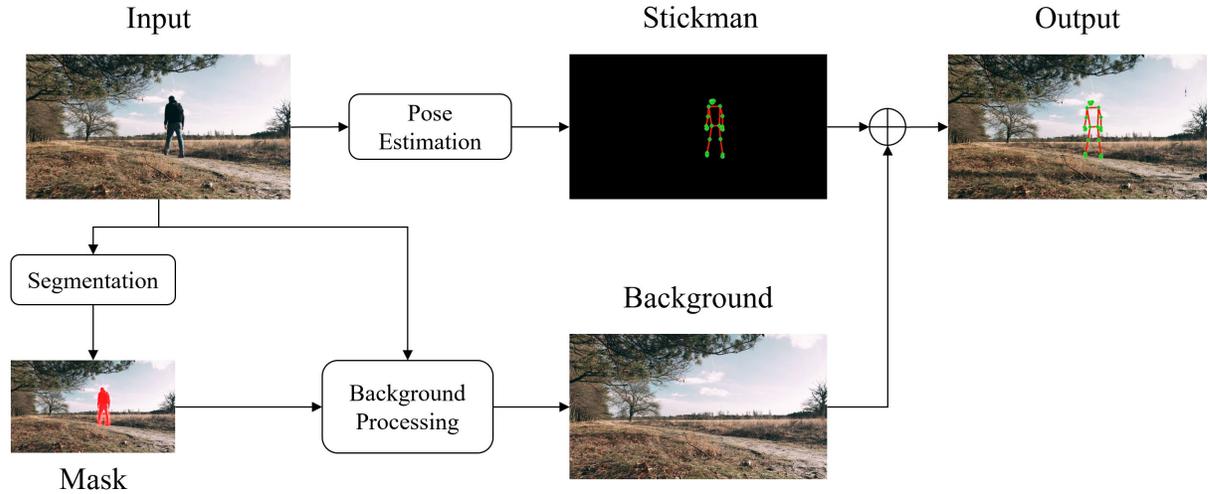

Figure 2. Anonymization Process Overview

### 2.3.3. Anonymization with Pose Estimation Algorithm

Umar Asif et al. proposed a privacy-preserving fall detection algorithm in 2020 [46]. Their algorithm performs human pose estimation on camera footage to extract a skeleton frame, and then adjusts the posture of a virtual avatar to mimic the same posture as the skeleton. Their research is the most similar to ours, but it has the limitation that all surrounding background information captured by the camera is lost, and only the virtual character, resembling a game character, is displayed against an empty background.

Our research also employs human pose estimation as the main technology for anonymization, allowing us to perform hazard detection simultaneously with anonymization. By applying our proposed technology to home IoT cameras, it would be possible to take quick actions in case of falls involving the elderly or patients. Additionally, our method preserves the background image, enabling fast and accurate transmission of information about the surrounding environment.

## 3. Methodology

To achieve anonymization, we analyzed the scene to estimate the pose of individuals and removed them to generate a background image without people. By overlaying the pose estimation results on the background, we created anonymized videos where all identifiable information such as facial features, skin color, gender, body build, clothing, and accessories is removed. Fig. 2 illustrates the whole process.

### 3.1. Background Processing to Remove Human

To remove individuals from the video and leave only the background, we employed a method that does not use generative AI. Although using the latest generative AI APIs can easily remove individuals and synthesize background images, inputting all scenes captured by the camera into the generative AI would require more than 30 API calls per second per camera. To avoid high latency and costs, we used a low-computation algorithm that can run on a local machine.

#### 3.1.1. Strategy

We utilized semantic segmentation to remove individuals from the background without using generative AI. Since the video is recorded with a stationary camera, it is possible to extract the area behind the individual by analyzing accumulated frames in the video, provided that the individual moves sufficiently within a wide area. To reduce computational load, we propose a method that analyzes the minimum number of frames necessary to extract a clean background.

#### 3.1.2. Frame Extraction at Intervals

We extracted frames from the video at regular intervals, including the first frame. Including the first frame is essential because if the initial frame does not contain any individuals, the background extraction process can be completed with just one computation.

When individuals move rapidly within the video, using shorter intervals can expedite the background extraction process. Conversely, if individuals are stationary or move slowly, wider intervals can be used, allowing the areas occupied by the individuals to remain as black silhouettes, thereby achieving quick anonymization.

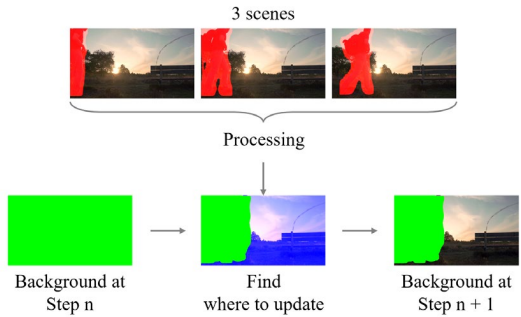

Figure 3. Background Update Rule

### 3.1.3. Semantic Segmentation of Human Body

Various methods can be utilized for semantic segmentation. Considering the video context, it is beneficial to reduce computational load using traditional optical flow-based semantic segmentation [47], or alternatively, using end-to-end algorithms like U-net [48] can be effective.

### 3.1.4. Background Update Strategy

The update rules for removing humans from the background are explained in Fig. 3 and Code 1. Initially, the background is an empty frame with all pixels set to -1. To update the background from step n to step n+1, it is essential to distinguish areas to be updated and those to remain 0. Update rules are applied to analyze the frames and update the background, as detailed in Fig. 4.

### 3.1.5. Update from Single Frame

The simplest method uses the segmentation mask from the most recent frame. Non-human classified pixels update the background where the corresponding coordinates are -1. This process continues until no -1 pixels remain or the proportion of -1 pixels falls below a threshold. A limitation of this method is early termination if the segmentation model makes false negative predictions, leaving too many empty spaces.

### 3.1.6. Update from Multiple Frames

Using segmentation masks from the most recent n frames improves accuracy. Fig. 3 and Code 1 illustrate this with three frames. The initial background is the same, but only pixels classified as non-human in all three frames update the background. This stricter criterion takes more time but avoids premature termination due to false negatives, ensuring a cleaner background. However, false positive predictions can extend the algorithm's duration.

```
Update Background
1   background = array(-1, (xdim, ydim, 3))
2   last_scenes = []
3   min_update_frames = 3
4   finish_treshold = 0.01
5
6   for frame in video_frames:
7       mask = model.predict(frame)
8       last_scenes.append(mask)
9
10      if len(last_scenes) < min_update_frames:
11          continue
12
13      diff = find_candidate(last_scenes)
14      background[diff] = frame[diff]
15
16      empty_area = count(background == -1, axis=2)
17      empty_rate = empty_area / (xdim * ydim)
18      if empty_rate < finish_treshold:
19          break
20
21      last_scenes = last_scenes[1:]
22
```

Code 1. Background Update Rule

### 3.1.7. Limitation

This update rule has limitations, as it can only be applied to videos recorded with a stationary camera. If the camera moves during recording, it is necessary to reinitialize the background from the point of movement and start the update process again.

## 3.2. of Human Body

To achieve final anonymization, we perform pose estimation on the video frames to create stickman annotations. These stickman figures are then drawn onto the background, effectively anonymizing the individuals in the video.

### 3.2.1. Human Pose Estimation Algorithm

By analyzing frames and performing human pose estimation, we generated stickman annotations. These stickman figures, which only display the positions of human joints without any skin, clothing, or facial information, were drawn on the erased background. Using a 3D inference algorithm, capable of depth prediction, is preferable to a 2D-scale inference algorithm for more accurate anonymization.

### 3.2.2. Anonymized Skeleton Generation

The inferred keypoints are represented as circles, and the connections between keypoints are drawn as lines, creating stickman annotations that are overlaid onto the background for anonymization.

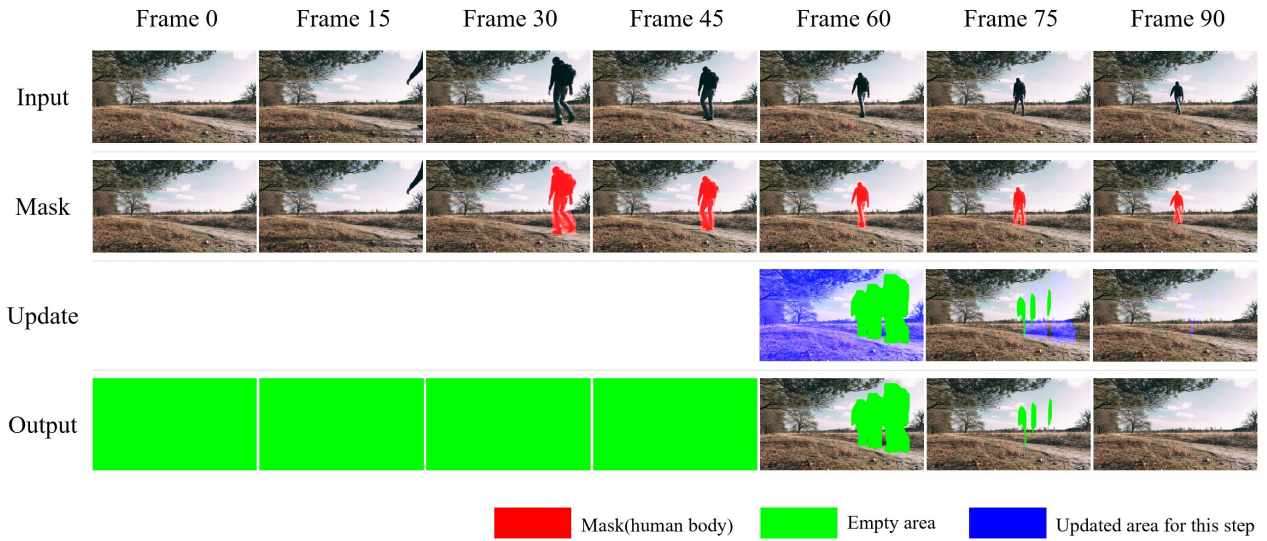

Figure 4. Summary of background update process to remove human body

### 3.3. Anonymized Scene Generation

By combining the background extraction algorithm and the pose estimation algorithm introduced earlier, we created a fully anonymized scene by removing people from the original images and overlaying stickman figures in their place. This process is illustrated in Fig. 1.

## 4. Experiments

### 4.1. Data Statement

Since we utilized a pretrained model, the data discussed here is used for method validation rather than training. All videos were sourced from the free stock video section on pixabay.com and are copyright-free. We collected a variety of videos shot with stationary cameras, including those where people move around, stay in one place, or move only slightly.

### 4.2. Experimental Setup

We conducted our research using an NVIDIA RTX 3080 GPU for the initial segmentation model. However, due to the low computational demand of MaskRCNN, we also used an Intel i9-13900K CPU for parallel experiments. For pose estimation using the BlazePose model, we exclusively utilized a CPU since the pretrained model by Google is designed for CPU operations.

We employed Python and Torch for neural network computations and OpenCV for processing mp4 videos.

### 4.3. Video Processing Workflow

#### 4.3.1. Video With Stationary Person

We extracted all frames through iterations for pose estimation. For background processing, we utilized frames extracted at intervals of 15 frames, including the initial frame. This interval corresponds to a 0.5-second gap in a 30fps video.

#### 4.3.2. Semantic Segmentation

We employed the Mask R-CNN [49] with a ResNet-50-FPN backbone model [50], pretrained on the COCO V1 dataset [51]. The model weights are 169.8MB, and it has 44.4 million parameters, making it relatively small. The COCO dataset consists of 80 categories, and the Mask R-CNN model used in our study is generalized to this dataset. Therefore, if a dataset specifically focused on human segmentation were used, and the model was trained from scratch to distinguish only two classes (background and human), it could be implemented with much smaller models.

#### 4.3.3. Background Update

To achieve the updates introduced in Fig. 3 and Code 1, we set a buffer to analyze three consecutive scenes, allowing for a single update. Given the occasional false negatives from the pretrained Mask R-CNN's human segmentation, we implemented additional algorithms to address these misses. Even after background updates were deemed complete, the algorithm continued updates if a human object was detected within three iterations. This method ensured the removal of undetected human areas by superimposing weights between the existing and new background update areas. This approach produces a translucent black shadow effect when individuals remain stationary.

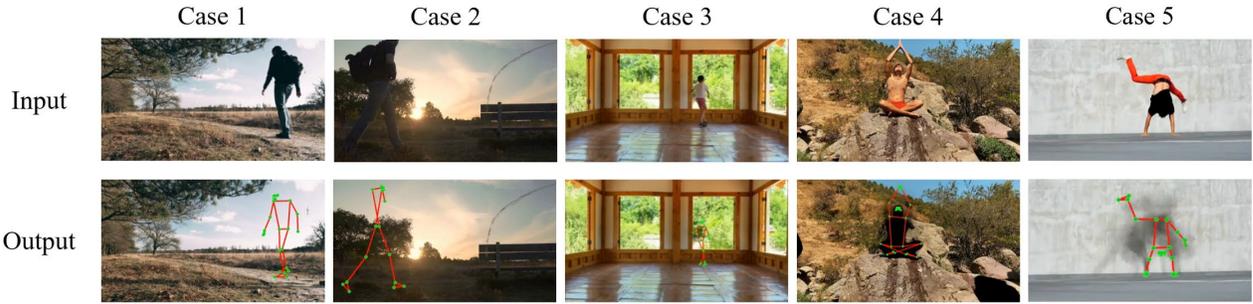

Figure 5. Processing results

### 4.3.4. Pose Estimation

We used the BlazePose algorithm [41] for human pose estimation. BlazePose is a lightweight convolutional neural network model designed for real-time inference on mobile devices. It can trace 33 body keypoints, including face detection, and predict 3-dimensional coordinates for all keypoints. We obtained the pretrained BlazePose model released by Google.

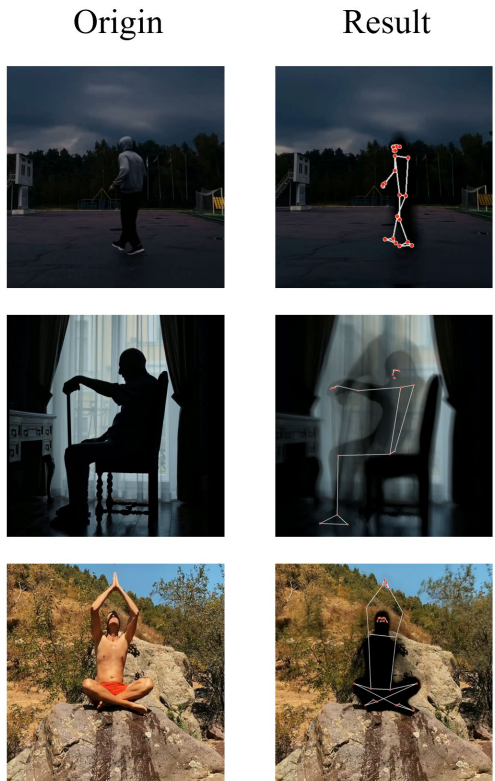

Figure 6. Anonymization of Video with Stationary Person

### 4.3.5. Anonymized Scene Generation

By compositing stickman annotations onto the processed background, we generated anonymized scenes. The resulting videos have the same resolution and frame rate as the original videos.

## 4.4. Evaluation

Most anonymization techniques use personality identification algorithms for performance evaluation. By inputting both original and processed videos into an identification model and statistically analyzing the difference in identification accuracy between the two groups, the performance of the anonymization algorithm can be quantitatively assessed.

However, our algorithm erases the entire human body, making it impossible to perform identification algorithm accuracy tests. Models that detect faces and perform identification based on cropped face images cannot recognize the presence of individuals in our method's output. Therefore, quantitative evaluation criteria are not applicable in this study.

## 5. Results and Discussion

## 5.1. Result Overview

We applied our method to several videos, and the resulting images are displayed in Fig. 5. Our approach successfully achieved full-body anonymization in all videos. None of the output images in Fig. 5 allow for the identification of individuals. Depending on the type of video, the background is either cleanly removed or a silhouette of the person remains. This phenomenon is determined by the amount of movement within the video. We will explain this by categorizing the videos into two types.

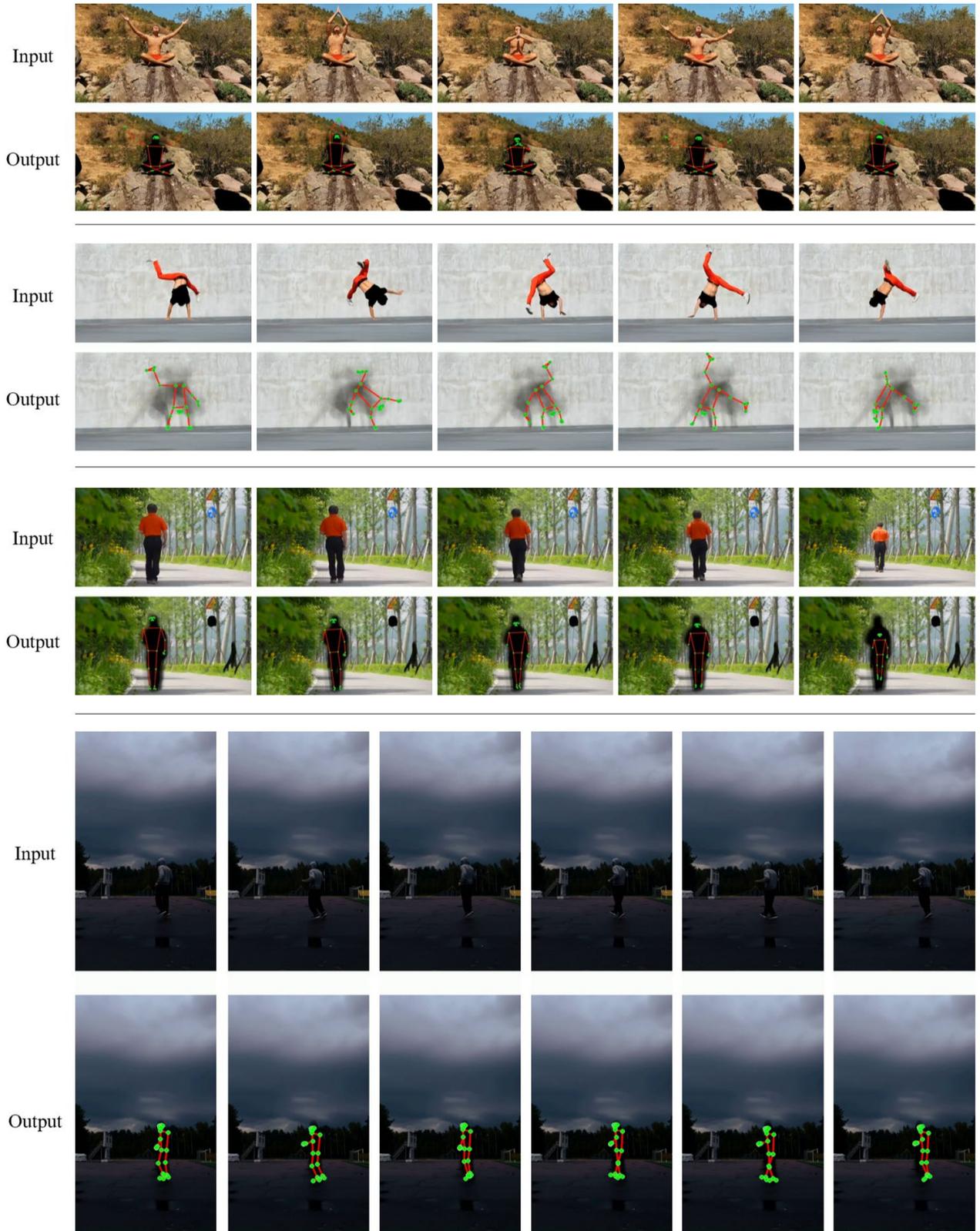

Figure 7. Anonymization of Videos with Stationary Person

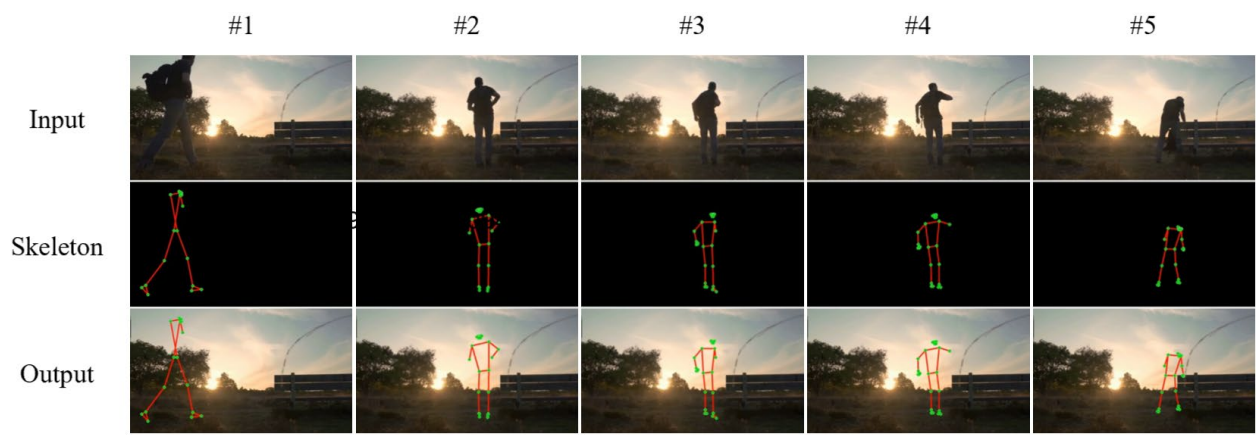
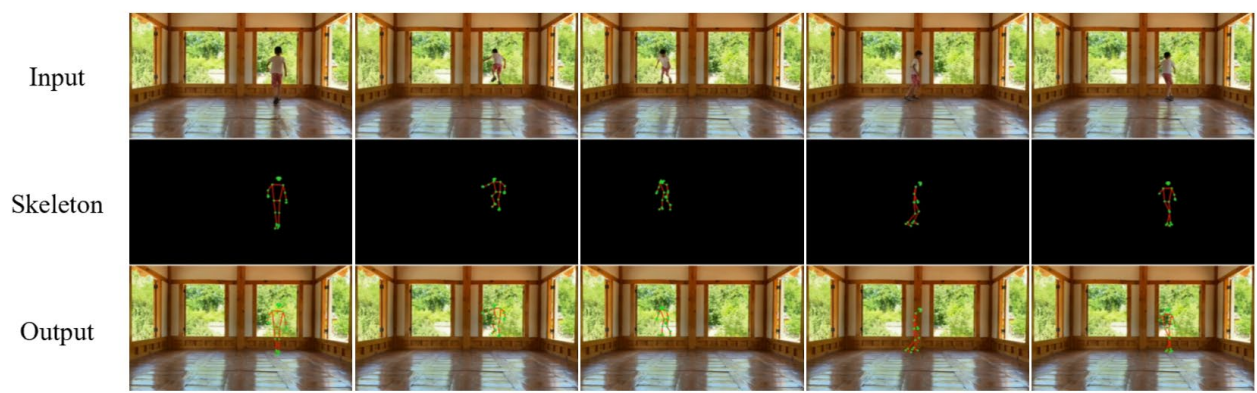
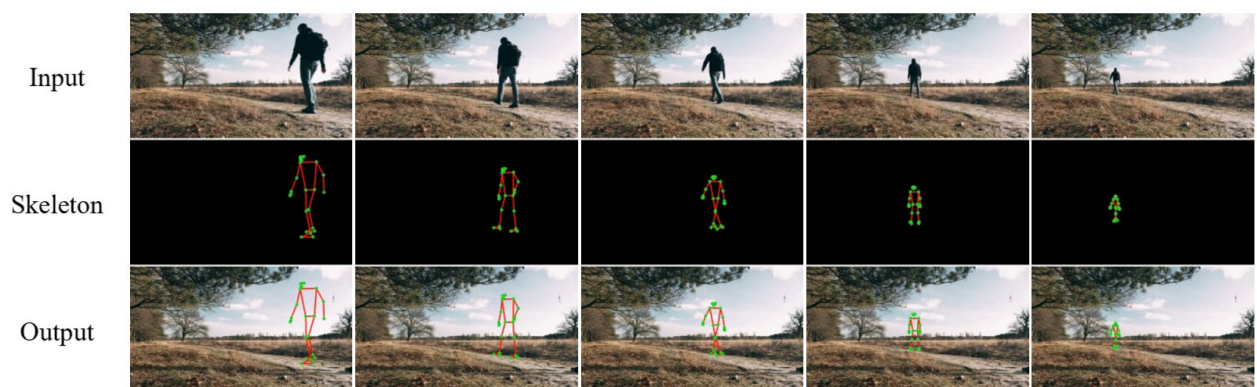

Figure 8. Anonymization of Videos with Moving Person

## 5.2. Video With Stationary Person

### 5.2.1. Result

In cases where there is minimal movement within the frame, the area where a person remains will be displayed as a black silhouette. Since our method does not employ generative AI, it does not fill in the erased areas with new data. Therefore, in instances of low movement, as shown in Fig. 6 and Fig. 7, the silhouette of the person will be visible, with the anonymized skeleton superimposed on it.

### 5.2.2. Discussion

If the movement of the subjects in the video is insufficient, residual images of the subjects may remain in the background, as shown in Fig. 6. Nevertheless, full body anonymization can still be achieved in such cases.

## 5.3. Video With Moving Person

### 5.3.1. Result

In videos where the person is moving, the background is updated in frames where objects that were previously obscured by the person become visible again as the person moves. As a result, a clear background can be obtained, as shown in Fig. 8. All identifiable information about the person is removed, achieving perfect anonymization.

### 5.3.2. Discussion

The clearest backgrounds are obtained in videos of moving people recorded by stationary cameras. This scenario appears to be optimal for applying our method. In the current industrial landscape, stationary camera devices such as CCTV and home IoT cameras are widely used in almost all locations. Therefore, despite the limitations posed by camera movement or the stationary state of individuals, our algorithm demonstrates high utility and applicability.

## 5.4. Model Complexity Comparison

Table 1 compares the number of parameters used by recent algorithms and our model. Our method achieves full-body anonymization using 44 million parameters. In contrast, DeepPrivacy [5] uses 47 million parameters to anonymize only faces. DeepPrivacy v2 [24] can anonymize full bodies but requires 129 million parameters.

Since most current anonymization technologies employ Generative Adversarial Networks (GANs), the parameters of the Discriminator also affect the computational speed during the training process. Furthermore, as end-to-end anonymization is pursued, the minimum number of model parameters significantly impacts the overall algorithm performance. Our algorithm simplifies the background removal process, allowing us to achieve anonymization with fewer parameters.

| Algorithm | Number of Parameters | Anonymization Target |
|---|---|---|
| **Ours** | **44M** | **Full-body** |
| DeepPrivacy [5] | 47M | Face |
| DeepPrivacy 2 [24] | 129M | Full-body |

Table 1. Model Comparison

## 5.5. Suggestions for Future Research

### 5.5.1. Applying Hazard Detection Algorithms

Our method employs human keypoint position estimation for anonymization, suggesting potential industrial applications by integrating hazard detection algorithms based on human posture estimation, such as fall detection algorithms. Implementing our method in industrial fields' CCTV systems, home IoT cameras, and hospitals could significantly enhance early emergency response and treatment, benefiting many individuals by providing timely and appropriate medical interventions.

### 5.5.2. Virtual Human Body Generation

By overlaying virtual images on skeletons and joints which are processed by human pose estimation algorithm, a virtual human body object can be generated. Displaying a skeleton object on an erased background can achieve anonymization, but overlaying a new virtual human image on this skeleton can open up various industrial applications.

## 6. Conclusion

Many researchers have developed techniques using generative AI to replace individuals in videos with entirely different people for anonymization. However, techniques that only alter the face fail to conceal personal information such as skin color, clothing, and accessories, leaving identifiable details. Previous attempts to replace both faces and clothing still struggle to obscure height, build, and posture.

We introduced a method that removes individuals from the background and overlays unidentifiable stickman annotations. This approach not only achieves the goal of anonymization but also allows for hazard detection by inferring the 3D coordinates of body joints, making it suitable for applications in home IoT services. Additionally, this technique can be utilized to overlay animated characters on the skeletons for various purposes, extending its use beyond anonymization. We expect this technology to be valuable in diverse fields.

## Acknowledgement

We extend our deepest gratitude to the creators who generously upload high-quality videos to the Pixabay Stock Video service. Your contributions have been invaluable in completing our research.

## References

[1] OECD, Privacy and data protection, https://www.oecd.org/en/topics/policy-issues/privacy-and-data-protection.html
[2] Rosberg, F., Aksoy, E. E., et al. (2023). FIVA: Facial Image and Video Anonymization and Anonymization Defense. In Proceedings of the IEEE/CVF International Conference on Computer Vision (pp. 362-371).
[3] Ren, Z., Lee, Y. J., and Ryoo, M. S. (2018). Learning to anonymize faces for privacy preserving action detection. In Proceedings of the european conference on computer vision (ECCV) (pp. 620-636).
[4] Lin, J., Li, Y., and Yang, G. (2021). FPGAN: Face de-identification method with generative adversarial networks for social robots. Neural Networks, 133, 132-147.
[5] Hukkelås, H., Mester, R., and Lindseth, F. (2019, October). Deepprivacy: A generative adversarial network for face anonymization. In International symposium on visual computing (pp. 565-578). Cham: Springer International Publishing.
[6] Hellmann, F., Mertes, S., et al. (2024). Ganonymization: A gan-based face anonymization framework for preserving emotional expressions. ACM Transactions on Multimedia Computing, Communications and Applications.
[7] Maximov, M., Elezi, I., & Leal-Taixé, L. (2020). Ciagan: Conditional identity anonymization generative adversarial networks. In Proceedings of the IEEE/CVF conference on computer vision and pattern recognition (pp. 5447-5456).
[8] Yang, K., Yau, J. H., Fei-Fei, L., Deng, J., & Russakovsky, O. (2022, June). A study of face obfuscation in imagenet. In International Conference on Machine Learning (pp. 25313-25330). PMLR.
[9] Redmon, J., Divvala, S., Girshick, R., & Farhadi, A. (2016). You only look once: Unified, real-time object detection. In Proceedings of the IEEE conference on computer vision and pattern recognition (pp. 779-788).
[10] Obaida, T., Hassan, N. F., & Jamil, A. S. (2022). Comparative of Viola-Jones and YOLO v3 for Face Detection in Real time. Iraqi Journal Of Computers, Communications, Control And Systems Engineering, 22(2), 63-72.
[11] Pebrianto, W., Mudjirahardjo, P., & Pramono, S. H. (2022, August). YOLO method analysis and comparison for real-time human face detection. In 2022 11th Electrical Power, Electronics, Communications, Controls and Informatics Seminar (EECCIS) (pp. 333-338). IEEE.
[12] Hasan, M. A., & Lazem, A. H. (2023). Facial Human Emotion Recognition by Using YOLO Faces Detection Algorithm. Central Asian Journal of Mathematical Theory And Computer Sciences, 4(10), 1-11.
[13] Bhambani, K., Jain, T., & Sultanpure, K. A. (2020, October). Real-time face mask and social distancing violation detection system using yolo. In 2020 IEEE Bangalore Humanitarian Technology Conference (B-HTC) (pp. 1-6). IEEE.
[14] Kale, Y. V., Shetty, A. U., Patil, Y. A., Patil, R. A., & Medhane, D. V. (2021, December). Object detection and face recognition using yolo and inception model. In International Conference on Advanced Network Technologies and Intelligent Computing (pp. 274-287). Cham: Springer International Publishing.
[15] Prayogo, B. P., Mulyana, E., & Hermawan, W. (2023, July). A Novel Approach for Face Recognition: YOLO-Based Face Detection and Facenet. In 2023 9th International Conference on Wireless and Telematics (ICWT) (pp. 1-6). IEEE.
[16] Yu, Z., Huang, H., Chen, W., Su, Y., Liu, Y., & Wang, X. (2024). Yolo-facev2: A scale and occlusion aware face detector. Pattern Recognition, 155, 110714.
[17] Real-Time Human Ear Detection Based on the Joint of Yolo and RetinaFace
[18] Aung, H., Bobkov, A. V., & Tun, N. L. (2021, May). Face detection in real time live video using yolo algorithm based on Vgg16 convolutional neural network. In 2021 International conference on industrial engineering, applications and manufacturing (ICIEAM) (pp. 697-702). IEEE.
[19] Garg, D., Goel, P., Pandya, S., Ganatra, A., & Kotecha, K. (2018, November). A deep learning approach for face detection using YOLO. In 2018 IEEE Punecon (pp. 1-4). IEEE.
[20] Goodfellow, I., Pouget-Abadie, J., Mirza, M., Xu, B., Warde-Farley, D., Ozair, S., ... & Bengio, Y. (2014). Generative adversarial nets. Advances in neural information processing systems, 27.
[21] Klemp, M., Rösch, K., Wagner, R., Quehl, J., & Lauer, M. (2023). LDFA: Latent diffusion face anonymization for self-driving applications. In Proceedings of the IEEE/CVF Conference on Computer Vision and Pattern Recognition (pp. 3199-3205).
[22] Lee, H. Y., Kwak, J. M., Ban, B., Na, S. J., Lee, S. R., & Lee, H. K. (2017, October). GAN-D: Generative adversarial networks for image deconvolution. In 2017 International Conference on Information and Communication Technology Convergence (ICTC) (pp. 132-137). IEEE.
[23] Byunghyun Ban. (2018). Cardiac CT image segmentation with adversarial networks., Masters Theses, Korea Advanced Institute of Science and Technology, Daejeon.
[24] Hukkelås, H., & Lindseth, F. (2023). Deepprivacy2: Towards realistic full-body anonymization. In Proceedings of the IEEE/CVF winter conference on applications of computer vision (pp. 1329-1338).
[25] Hukkelås, H., Smebye, M., Mester, R., & Lindseth, F. (2023). Realistic full-body anonymization with surface-guided GANs. In Proceedings of the IEEE/CVF Winter conference on Applications of Computer Vision (pp. 1430-1440).
[26] Zarif, S., Faye, I., & Rohaya, D. (2013, March). Static object removal from video scene using local similarity. In 2013

IEEE 9th international colloquium on signal processing and its applications (pp. 54-57). IEEE.

[27] Ban, B., Ryu, D., & Hwang, S. W. (2023, October). CongNaMul: A Dataset for Advanced Image Processing of Soybean Sprouts. In 2023 14th International Conference on Information and Communication Technology Convergence (ICTC) (pp. 1892-1897). IEEE.

[28] Chen, H. W., & McGurr, M. (2014, August). Improved color and intensity patch segmentation for human full-body and body-parts detection and tracking. In 2014 11th IEEE International Conference on Advanced Video and Signal Based Surveillance (AVSS) (pp. 361-368). IEEE.

[29] Shaker, A. S. (2020). Detection and segmentation of osteoporosis in human body using recurrent neural network. International Journal of Advanced Science and Technology, 29(02), 1055-1066.

[30] Zhang, S. H., Li, R., Dong, X., Rosin, P., Cai, Z., Han, X., ... & Hu, S. M. (2019). Pose2seg: Detection free human instance segmentation. In Proceedings of the IEEE/CVF conference on computer vision and pattern recognition (pp. 889-898).

[31] Lin, C. Y., Xie, H. X., & Zheng, H. (2019). PedJointNet: Joint head-shoulder and full body deep network for pedestrian detection. IEEE Access, 7, 47687-47697.

[32] Caputo, S., Castellano, G., Greco, F., Mencar, C., Petti, N., & Vessio, G. (2021, December). Human detection in drone images using yolo for search-and-rescue operations. In International Conference of the Italian Association for Artificial Intelligence (pp. 326-337). Cham: Springer International Publishing.

[33] Alruwaili, M., Siddiqi, M. H., Atta, M. N., & Arif, M. (2024). Deep learning and ubiquitous systems for disabled people detection using YOLO models. Computers in Human Behavior, 154, 108150.

[34] Granados, M., Tompkin, J., Kim, K., Grau, O., Kautz, J., & Theobalt, C. (2012, May). How not to be seen—object removal from videos of crowded scenes. In Computer Graphics Forum (Vol. 31, No. 2pt1, pp. 219-228). Oxford, UK: Blackwell Publishing Ltd.

[35] Fu, Q., Guan, Y., & Yang, Y. (2018). Image inpainting and object removal with deep convolutional gan. Technical report. http://stanford. edu/class/ee367/Winter2018.

[36] Dantone, M., Gall, J., Leistner, C., & Van Gool, L. (2013). Human pose estimation using body parts dependent joint regressors. In Proceedings of the IEEE conference on computer vision and pattern recognition (pp. 3041-3048).

[37] Sun, M., Kohli, P., & Shotton, J. (2012, June). Conditional regression forests for human pose estimation. In 2012 IEEE Conference on Computer Vision and Pattern Recognition (pp. 3394-3401). IEEE.

[38] Zhang, H., Ouyang, H., Liu, S., Qi, X., Shen, X., Yang, R., & Jia, J. (2019). Human pose estimation with spatial contextual information. arXiv preprint arXiv:1901.01760.

[39] Toshev, A., & Szegedy, C. (2014). Deeppose: Human pose estimation via deep neural networks. In Proceedings of the IEEE conference on computer vision and pattern recognition (pp. 1653-1660).

[40] Véges, M., & Lőrincz, A. (2019, July). Absolute human pose estimation with depth prediction network. In 2019 International Joint Conference on Neural Networks (IJCNN) (pp. 1-7). IEEE.

[41] Bazarevsky, V., Grishchenko, I., Raveendran, K., Zhu, T., Zhang, F., & Grundmann, M. (2020). Blazepose: On-device real-time body pose tracking. arXiv preprint arXiv:2006.10204.

[42] Balasubramanian, C. R. (2020). Fall risk monitoring scheme based on human posture estimation using Transfer learning (Doctoral dissertation, Dublin, National College of Ireland).

[43] Adhikari, K., Bouchachia, H., & Nait-Charif, H. (2019). Deep learning based fall detection using simplified human posture. Int. J. Comput. Syst. Eng, 13(5), 255-260.

[44] Shi, L., Xue, H., Meng, C., Gao, Y., & Wei, L. (2023, July). DSC-OpenPose: A Fall Detection Algorithm Based on Posture Estimation Model. In International Conference on Intelligent Computing (pp. 263-276). Singapore: Springer Nature Singapore.

[45] Lee, S., Choi, W., Park, M., Jeon, Y., Tran, Q., & Park, S. (2023). Chapter Deep Learning-Based Pose Estimation for Identifying Potential Fall Hazards of Construction Worker.

[46] Asif, U., Mashford, B., Von Cavallar, S., Yohanandan, S., Roy, S., Tang, J., & Harrer, S. (2020, April). Privacy preserving human fall detection using video data. In Machine Learning for Health Workshop (pp. 39-51). PMLR.

[47] Grundmann, M., Kwatra, V., Han, M., & Essa, I. (2010, June). Efficient hierarchical graph-based video segmentation. In 2010 ieee computer society conference on computer vision and pattern recognition (pp. 2141-2148). IEEE.

[48] Ronneberger, O., Fischer, P., & Brox, T. (2015). U-net: Convolutional networks for biomedical image segmentation. In Medical image computing and computer-assisted intervention–MICCAI 2015: 18th international conference, Munich, Germany, October 5-9, 2015, proceedings, part III 18 (pp. 234-241). Springer International Publishing.

[49] He, K., Gkioxari, G., Dollár, P., & Girshick, R. (2017). Mask r-cnn. In Proceedings of the IEEE international conference on computer vision (pp. 2961-2969).

[50] Lin, T. Y., Dollár, P., Girshick, R., He, K., Hariharan, B., & Belongie, S. (2017). Feature pyramid networks for object detection. In Proceedings of the IEEE conference on computer vision and pattern recognition (pp. 2117-2125).

[51] Lin, T. Y., Maire, M., Belongie, S., Hays, J., Perona, P., Ramanan, D., ... & Zitnick, C. L. (2014). Microsoft coco: Common objects in context. In Computer Vision–ECCV 2014: 13th European Conference, Zurich, Switzerland, September 6-12, 2014, Proceedings, Part V 13 (pp. 740-755). Springer International Publishing.